\documentclass[conference]{IEEEtran}
\IEEEoverridecommandlockouts
\usepackage{cite}
\usepackage{amsmath,amssymb,amsfonts}
\usepackage{algorithmic}
\usepackage{subfigure}
\usepackage{graphicx}
\usepackage{textcomp}
\usepackage{xcolor}
\usepackage{url}
\usepackage{multirow}
\usepackage{booktabs}
\def\BibTeX{{\rm B\kern-.05em{\sc i\kern-.025em b}\kern-.08em
    T\kern-.1667em\lower.7ex\hbox{E}\kern-.125emX}}

\usepackage{xparse} 
\NewDocumentCommand{\yixin}{ mO{} }{\textcolor{blue}{\textsuperscript{\textit{Yixin}}\textsf{\textbf{\small[#1]}}}}

\NewDocumentCommand{\jun}{ mO{} }{\textcolor{blue}{\textsuperscript{\textit{Jun}}\textsf{\textbf{\small[#1]}}}}

\NewDocumentCommand{\method}{ mO{} }{#1}

\begin{document}

\title{Improving Neural Relation Extraction with Implicit Mutual Relations\\
}

\author{\IEEEauthorblockN{Jun Kuang\IEEEauthorrefmark{1}, Yixin Cao\IEEEauthorrefmark{2}, Jianbing Zheng\IEEEauthorrefmark{1}, Xiangnan He\IEEEauthorrefmark{3}, Ming Gao\IEEEauthorrefmark{1}\IEEEauthorrefmark{4}, Aoying Zhou\IEEEauthorrefmark{1}} \\
\IEEEauthorblockA{
\IEEEauthorrefmark{1}\textit{School of Data Science and Engineering, East China Normal University, Shanghai, China}
}

\IEEEauthorblockA{
\IEEEauthorrefmark{2}\textit{School of Computing, National University of Singapore, Singapore}
}

\IEEEauthorblockA{
\IEEEauthorrefmark{3}\textit{School of Information Science and Technology, University of Science and Technology of China, Hefei, China}
}
\IEEEauthorblockA{
\IEEEauthorrefmark{4}\textit{KLATASDS-MOE, School of Statistics, East China Normal University, Shanghai, China}
} 

}
\maketitle

\begin{abstract}
Relation extraction (RE) aims at extracting the relation between two entities from the text corpora. It is a crucial task for Knowledge Graph (KG) construction. Most existing methods predict the relation between an entity pair by learning the relation from the training sentences, which contain the targeted entity pair. In contrast to existing distant supervision approaches that suffer from insufficient training corpora to extract relations, our proposal of mining implicit mutual relation from the massive unlabeled corpora transfers the semantic information of entity pairs into the RE model, which is more expressive and semantically plausible. After constructing an entity proximity graph based on the implicit mutual relations, we preserve the semantic relations of entity pairs via embedding each vertex of the graph into a low-dimensional space. As a result, we can easily and flexibly integrate the implicit mutual relations and other entity information, such as entity types, into the existing RE methods.  

Our experimental results on a New York Times and another Google Distant Supervision datasets suggest that our proposed neural RE framework provides a promising improvement for the RE task, and significantly outperforms the state-of-the-art methods. Moreover, the component for mining implicit mutual relations is so flexible that can help to improve the performance of both CNN-based and RNN-based RE models significant.
\end{abstract}


\begin{IEEEkeywords}
Relation extraction, implicit mutual relations, unlabeled data, entity information
\end{IEEEkeywords} 
\section{INTRODUCTION}
Recently, we have witnessed an ocean of Knowledge Graphs (KGs), such as DBpedia~\cite{dbpedia}, FreeBase~\cite{freebase}, and YAGO~\cite{yago}, which has been successfully applied in a host of tasks, including question answering~\cite{kbqa}, search engine~\cite{searchengine} and chat robot~\cite{chatbot}. These KGs are far from complete. Thus, it attracts much attention to extract factual triplet from plain text for KG completion, e.g., (\textbf{Obama}, \emph{born}, \textbf{Hawaii}), which involves two sub-tasks of entity linking~\cite{shen2014entity} and relation extraction (RE)~\cite{bach2007review}.

As a paramount step, RE is typically regarded as a classification problem~\cite{Kambhatla}. Given two entities (e.g., \textbf{Obama} and \textbf{Hawaii}), RE aims at classifying them into pre-defined relation types (e.g., \emph{born}) based on the sentences involving the entity pair. It is nontrivial since the same relation may have various textual expressions, and meanwhile, different relations can also be described using the same words.

\begin{figure}[htb]

\centering
\begin{minipage}[t]{0.45\linewidth}
\subfigure[NYT]
{
\label{fig:nyt_statisic}
 \includegraphics[scale=0.28]{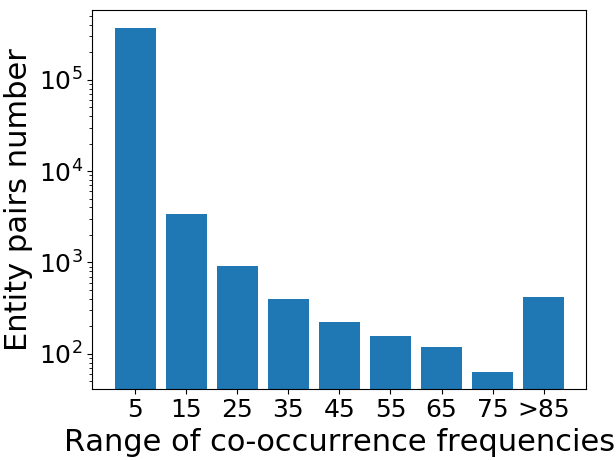}
}
\end{minipage}
\hspace*{\fill}
\centering
\begin{minipage}[t]{0.45\linewidth}
\subfigure[GDS]
{
    \label{fig:gds_statisic}
    \includegraphics[scale=0.28]{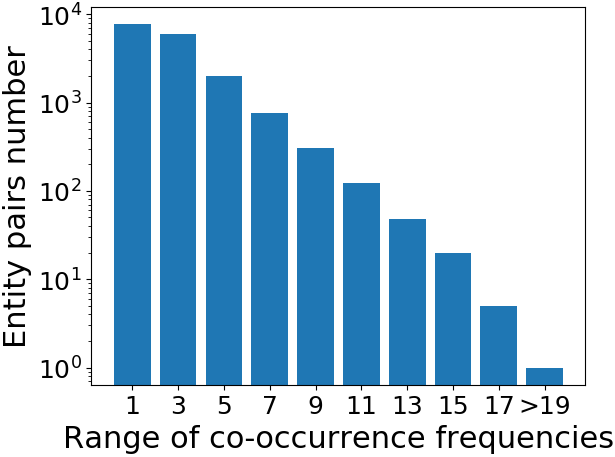}
}
\end{minipage}
\hspace*{\fill}
\caption{The number of entity pairs with different training data via distant supervision. The y-axis uses the log-scale.}
\label{fig:statistic}

\end{figure}

Existing RE approaches have achieved a great success based on deep neural network (NN)~\cite{cnn, wordatt}. They encode the texts via CNN~\cite{cnn} or RNN~\cite{wordatt} without feature engineering, then feed the hidden states into a softmax layer for classification. However, there are two issues arise from NN-based RE models:

\noindent\textbf{Insufficient Training Corpora} For satisfactory performance, these NN-based models require a large amount of training data, which is usually expensive to obtain. Alternatively, distant supervision is proposed to automatically extract sentences for training~\cite{ds}. It is under the assumption that if two entities $(head,tail)$ participate in a relation $r$, any sentence that contains $head$ and $tail$ might express that relation. However, there are still many infrequent entity pairs lacking sufficient training data due to the long-tailed distribution of frequencies of entity pairs. As illustrated in Figure~\ref{fig:statistic}, we count the number of entity pairs in log-scale with different range of co-occurrence frequencies in the dataset. The x-axis denotes the range of co-occurrence frequencies in the corresponding dataset. The y-axis denotes the number of entity pairs which co-occurrence frequencies are in the corresponding range. We can find that more than 90\% of the entity pairs in the GDS dataset have co-occurrence frequencies less than 10, and this situation becomes more severe in NYT dataset. 

\begin{table*}[t]
\caption{An example of the implicit mutual relations between entity pairs.}
\begin{center}
\begin{tabular}{|c|c|c|c|c|}
\hline
ID & \textbf{Entity pair}                             & \textbf{Sentences} & \textbf{Sentence example}                                                               & \textbf{Relation} \\ \hline
1  & (Stanford University, California)                & 2                  & ...and the \textbf{California}, ...learned from \textbf{Stanford University}...       & \color{red}hard to extract   \\ \hline
2  & (University of Washington, Seattle)              & 17                 & ...research at the \textbf{University of Washington} in \textbf{Seattle}...    & \textit{locatedIn}         \\ \hline
3  & (University of Southern California, Los Angeles) & 13                  & ...at the \textbf{University of Southern California} in \textbf{Los Angeles}... & \textit{locatedIn}         \\ \hline

4 &(Columbia University, New York City) & 24
& ...in \textbf{New York City}, ...graduated from \textbf{Columbia University}... & \textit{locatedIn} \\ \hline
\end{tabular}
\end{center}

\label{table:imeg}
\end{table*}

\noindent\textbf{Noisy Data} Although a large number of labeled data can be employed to train the RE algorithms with the help of distant supervision, the assumption sometimes is too strong and may introduce much noise. For example, the sentence ``Barack Obama visits Hawaii'' cannot express the relation \textit{born in}, but distant supervision would take it as ground truth. Existing methods usually alleviate the negative impact of noise by utilizing attention mechanism~\cite{senatt}. They select high-quality sentences by assigning them to higher weights and reduce the impact of noisy sentences through setting lower weights to them. However, we argue that the abandon of sentences may exacerbate the inadequate issue of training data.

By intuition, if different entity pairs are similar in semantic, they are more likely to have the same relation. 
For example, we illustrate four entity pairs in Table~\ref{table:imeg}, where they are similar in semantic, and all have the \textit{locatedIn} relation.
For target entity pair $ep_1 = $(\textbf{Stanford University}, \textbf{California}), if there are only two sentences in the distant supervision training dataset, its relation is not easily predicted due to the insufficient training instances and noisy data (e.g., the listed sentence in ID1 of Table~\ref{table:imeg} cannot express the \textit{locatedIn} relation of $ep_1$).
As illustrated in Figure~\ref{fig:statistic}, infrequent entity pairs are very common cases in the distant supervision training datasets.
Fortunately, all entity pairs, such as $ep_2$, $ep_3$ and $ep_4$, are helpful to predict the relation of the target entity pair $ep_1$.
Not limited to this, for target entity pair $ep_2 = $(\textbf{University of Washington}, \textbf{Seattle}), the semantic information of entity pair $ep_3$ = (\textbf{University of Southern California, Los Angeles}) is similar to the target entity pair $ep_2$, and therefore helpful to predict the relation of entity pair $ep_2$, vice versa.
In a word, all entity pairs with similar semantic are helpful to extract relation to each other.

However, most of the approaches cannot capture the similar semantic from the training dataset since only sentences, which contain the target entity pair, are employed to train the RE model.
Many existing works, such as Word2vec~\cite{word2vec}, GloVe~\cite{glove}, and BERT~\cite{bert}, etc., can extract the semantic information of words from the unlabeled corpora, rather than entities.
In contrast to extracting semantic information of words, we aim at mining the semantic information of entities to furthermore improve the performance of RE model.

To capture the mutual semantic relation between entities, we construct an entity proximity graph based on the unlabeled corpora, and further employ an embedding-based approach to learn a low-dimensional representation of each entity. 
As a result, the relations between entities are implicitly represented in the low-dimensional and semantic space, where entities with the similar semantic are closed in the space. 
Thus, we call the relations as \textbf{implicit mutual relations} of entities.
In particular, the component for mining implicit mutual relations can be seamlessly and flexibly integrated into the other RE models, such as CNN-based and RNN-based approaches.

In addition, the relation \textit{place of birth} must be a relation between two entities whose types are \textit{Location} and \textit{Person}. 
Entity types are therefore helpful to predict the relation of a target entity pair.
Thus, except for implicit mutual relations, we also integrate the distant supervision training data and the other entity information, such as 
entity types, to further improve the RE model. 
%
The main contributions of this paper are as follows: 
\begin{itemize}

    \item We propose to utilize implicit mutual relations between entity pairs to improve the RE task, and we mine such mutual relations from the easily available unlabeled data.
    
    \item We design a unified and flexible deep neural network framework, which ensembles the training corpora, entity types and implicit mutual relations, is proposed to extract relation from the plain text.

    \item We evaluate the proposed algorithm against baselines on two datasets. Experimental results illustrate the promising performance, and indicate that the implicit mutual relations are rewarding to improve the performance of both CNN-based and RNN-based RE models.
\end{itemize}

The rest of the paper is organized as follows. Section~\ref{sec:rw} covers the related works. In Section~\ref{sec:solution}, we formulate the problem formally, and provide our solution for RE. We report the promising experiment results on real-world datasets in Section~\ref{sec:exp}. Finally, we conclude the paper in Section~\ref{sec:conculsion}.

\section{RELATED WORK}~\label{sec:rw}

For extracting relations from the training corpora, supervised learning methods 
are the most effective~\cite{rereview}.

Especially, the neural network methods for relation extraction have also made a great progress in recent years. Socher et al.~\cite{socher} first parse the sentences and then use a recursive network to encode the sentences. Zeng et al.~\cite{cnn} propose a CNN-based model which can capture the lexical and sentence level features. Zeng et al.~\cite{pcnn} improve the CNN-based model by using the piecewise max pooling in the pooling layer of CNN.

Lots of works are focusing on improving the performances of the neural network methods, these works mainly start from the following aspects:

\begin{itemize}
    \item The neural encoder is adopted to extract various features from training corpora, such as syntax, semantics, etc~\cite{walked}\cite{miwa2016end}. The encoders which better capture and express the information can lead to better performance on relation extraction. Therefore, many works focus on improving the neural encoder to get more prominent relation extraction models.
    
    \item The neural network methods perform well for relation extraction. However, These methods require labor overhead for data annotation. As a result, the problem of lacking labeled data is more serious for large scale datasets. To address the issue, the distant supervision learning is proposed~\cite{ds}. The distant supervision learning is under the assumption that if an entity pair $(head,tail)$ has a relation $r$, any sentences that contain $head$ and $tail$ might express this relation. So labeled data can be obtained by aligning training corpora to KGs. However, the distant supervision will inevitably introduce the noise into the training data. Therefore, many works attempt to address how to alleviate the performance loss caused by noisy data~\cite{senatt}\cite{riedel}\cite{hoffmann}.
    
    \item Some works extract the relations of targeted entity pair only using the text which contains the entities in the target pair, while the others try to improve the relation extraction via mining the extra useful information, such as relation alias information~\cite{reside2018}, relation path~\cite{relationpath}, and entity description~\cite{ji2017distant}, etc. This extra information can be mined from various sources, including labeled and unlabeled data. The researchers integrate the extra information into relation extraction model as a supplementary, to enrich the information which relation extraction needs.
\end{itemize}


\subsection{Neural Encoder Improvement}

Some works design more sophisticated neural network encoders to improve the performance of relational extraction. Santos et al.~\cite{crcnn} use a convolutional neural network that performs relation extraction by ranking(CR-CNN). Nguyen et al.~\cite{nguyen} utilize multiple window sizes for CNN filters to obtain more features. Miwa et al.~\cite{miwa2016end} stack bidirectional tree-structured LSTM-RNNs on bidirectional sequential LSTM-RNNs to encode both word sequence and dependency tree substructure information.
Moreover, Christopoulou et al.~\cite{walked} encode multiple entity pairs in a sentence simultaneously to make use of the interaction among them. They place all the entities in a sentence as nodes in a full-connected entity graph, and encode them with a walk-based model on the entity graph. 

\subsection{Noise Mitigation}

To mitigate the noise in distant supervision learning, some works~\cite{riedel}\cite{hoffmann} utilize the multi-instance learning which allows different sentences to have at most one shared label. The multi-instance learning combines all relevant instances to determine the relation of the targeted entity pair, thereby alleviating the impact of wrong labeled instances. Surdeanu et al.~\cite{surdeanu} get rid of the restrict that different sentences can only share one label by utilizing a graphical model which can jointly model the multiple instances and multiple relations. 

With the development of the neural network, a technique called attention mechanism is proposed~\cite{attention}. The attention mechanism can let neural network models focus on the important training sentences. In the field of relation extraction, attention mechanism is widely used to mitigate the effects of noisy data~\cite{P1183-1194}. Existing attention approaches can be categorized into two groups: sentence-level attention and word-level attention. Sentence-level attention~\cite{senatt} aims at selecting the sentences w.r.t. the relational strength between the target entity pair. Similarly, word-level attention~\cite{wordatt} focuses on high-quality words to measure the target relation.  Furthermore, some works~\cite{D18-1243}~\cite{D18-1245}~\cite{wang2016relation} adopt the hierarchical attention which combines these two attention mechanisms, and further improves the performance of relation extraction.

Alternatively, reinforcement learning can also alleviate the effects of noisy data~\cite{reinforce}~\cite{Qin2018RobustDS}. The reinforcement learning methods mainly consist of two modules: a module is called instance selector to select the high-quality instances, and the other module is called relation classifier to make the prediction and provide rewards to the instance selector. The noisy data will be eliminated by the instance selector, that leads to a performance improvement.

Adversarial training is also a viable solution to address the noise problem. Wu et al.~\cite{adversarial} introduce adversarial training~\cite{goodfellow2014explaining} into the relation extraction task. They generate adversarial samples by first adding noise in the form of small perturbations to the original data, then encouraging the neural network to correctly classify both unmodified examples and perturbed ones to regularizing the relation extraction model. The regularized relation extraction model is more robusted and has higher generalization performance, so it can fight noise data very well. Furthermore, Qin et al.~\cite{qin2018dsgan} utilize the Generative Adaversarial Networks(GANs)~\cite{gans} to filter distant supervision training dataset and redistribute the false positive instances into the negative set.

\subsection{Extra Information supplementary}
The other direction to improve the performance of relation extraction model is to integrate more useful information into the existing approaches. This extra information is a good supplementary because this information cannot be extracted from the training corpora directly.

Some works attempt to introduce extra relation information.~\cite{reside2018} et al. utilize the relation alias information (e.g. \textit{founded} and \textit{co-founded} are aliases for the relation \textit{founderOfCompany}) to enhance the relation extraction. Zeng et al.~\cite{relationpath} construct the relation path between two entities that are not in the same sentence. 
Ji et al~\cite{ji2017distant} utilize the entity description information to supplement background knowledge. Liu et al.~\cite{entitytype} improve the relation extraction with entity type information. 
In addition, the semantic information~\cite{D18-1201} and part-of-speech tag~\cite{postag} are also good supplementary.

Although the additional information can improve the performance of relation extraction models, some of the information relies on high-quality sources which are expensive to collect. In our solution, we mine the implicit mutual relations between entity pairs from the available unlabeled data. 
In addition, the entity type information we used is also easily obtained via aligning the training corpora to KGs, which contain the entity type information. 


\section{METHODOLOGY}~\label{sec:solution}
\begin{figure*}[htp]
\centering
\includegraphics[width=15cm]{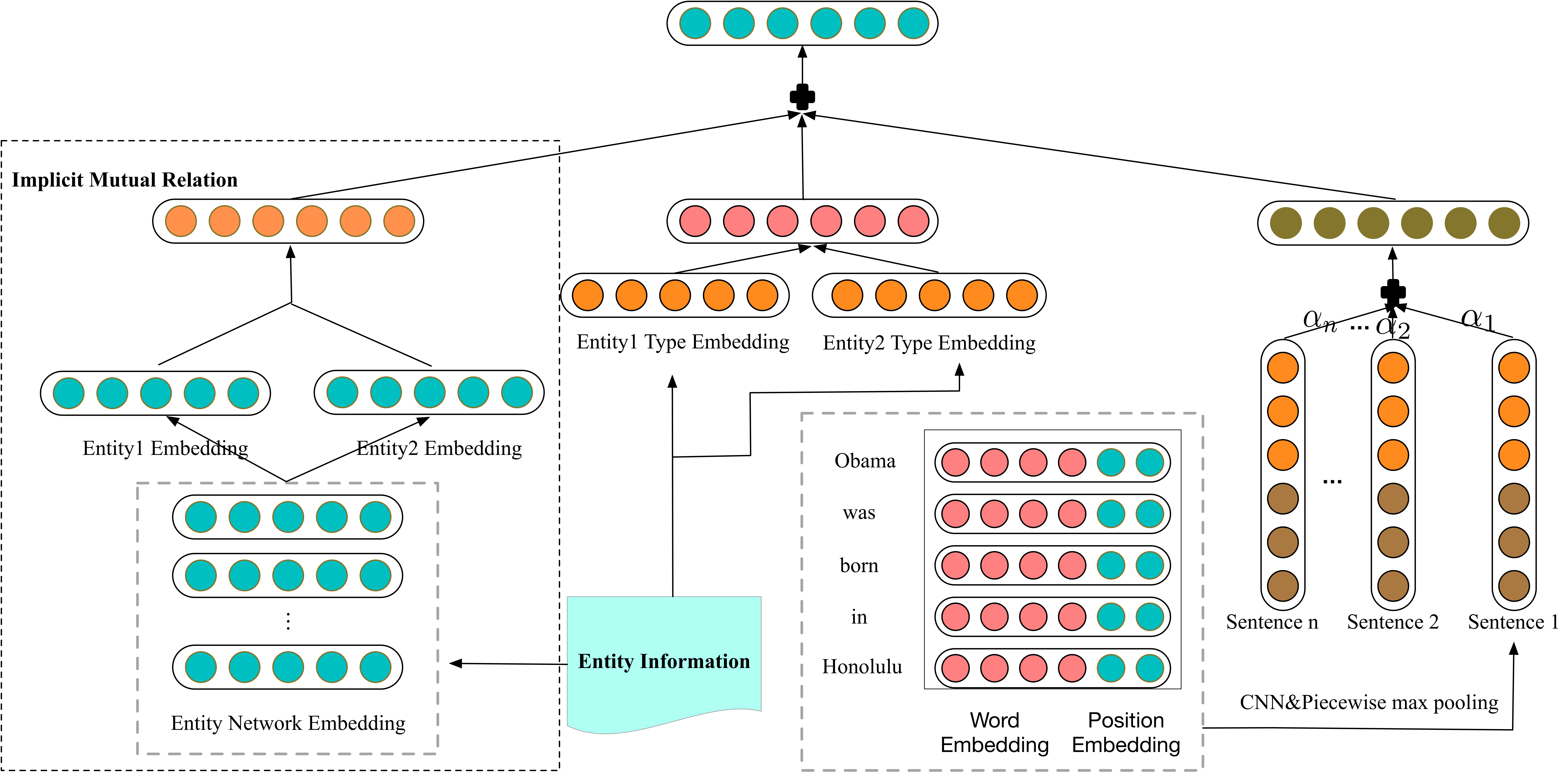}
\caption{Overview of our neural relation extraction framework.}
\label{fig:framework}
\end{figure*}

Given a target entity pair $(head, tail)$, and a set of training sentences $S = \{s_1,s_2,\cdots,s_n\}$, where each sentence $s_i$ contains the entities $head$ and $tail$. Our model aims at classifying the relation $r$ between entities $head$ and $tail$ by utilizing the sentences, the implicit mutual relations and the entity type information. As illustrated in Figure~\ref{fig:framework}, our proposed algorithm consists of four components:
\begin{itemize}
\item \textbf{Implicit Mutual Relations Modeling}: 
We construct an entity proximity graph to mine the entity pairs with high semantics proximity. In the graph, the semantics proximity can be defined as co-occurrence or similarity between entities in an external unlabeled corpora, rather than the training corpora. Thus, the graph can be constructed in an unsupervised manner. The entities with similar semantics have a similar topological structure in the entity proximity graph. Thus, the implicit mutual relation can be captured by the proximity graph. After vertex embedding, entities will project a low-dimensional space, where entities with similar semantic are closed in the embedding space. 

\item \textbf{Entity Type Embedding}:  The entity type is beneficial to filter impossible relations between two entities. For example, entities \textbf{Obama} and \textbf{Hawaii} are person and location, respectively. The relation between them is absolutely not \textit{childOf}. Thus, we first collect the types of corresponding entities from Freebase , and then embed them into a low-dimensional space. Then we can calculate a confidence score of each relation for the target entity pair by the entity type embedding. The confidence score of relation $r$ means the probability that there is a relation $r$ between the target entity pair.

\item \textbf{Piecewise CNN with Sentence-Level Attention}:  We use the PCNN to encode each sentence $s_i$ into $x_i$, then the sentences bag $S$ is encoded into $X = \{ x_1,x_2,\cdots,x_n\}$. To mitigate effect from the noisy sentence, a sentence-level attention is employed to focus the high quality sentences.

\item \textbf{Integrating Implicit Mutual Relation and Entity Type into RE Method}:  Finally, we integrate the entity types and implicit mutual relation into existing RE approaches. The implicit mutual relations, entity type embedding, and original RE model can calculate the confidence score of each relation separately. The confidence score means the probability that the target entity pair have the corresponding relation. We unify these confidence scores by a linear model and then get the probability that the target entity pair have the relation $r$.
\end{itemize}


\subsection{Implicit Mutual Relations Modeling}

There are three stages for implicit mutual relations modeling: (1) we construct an entity proximity graph based on the co-existing times of each entity pair; (2) then the entity representation is learned based on the entity proximity graph; (3) we model the implicit mutual relations by the entity representation. The details are shown as follow:

\subsubsection{Entity proximity graph construction}

\begin{figure}[htp]
\centering
\includegraphics[scale=0.5]{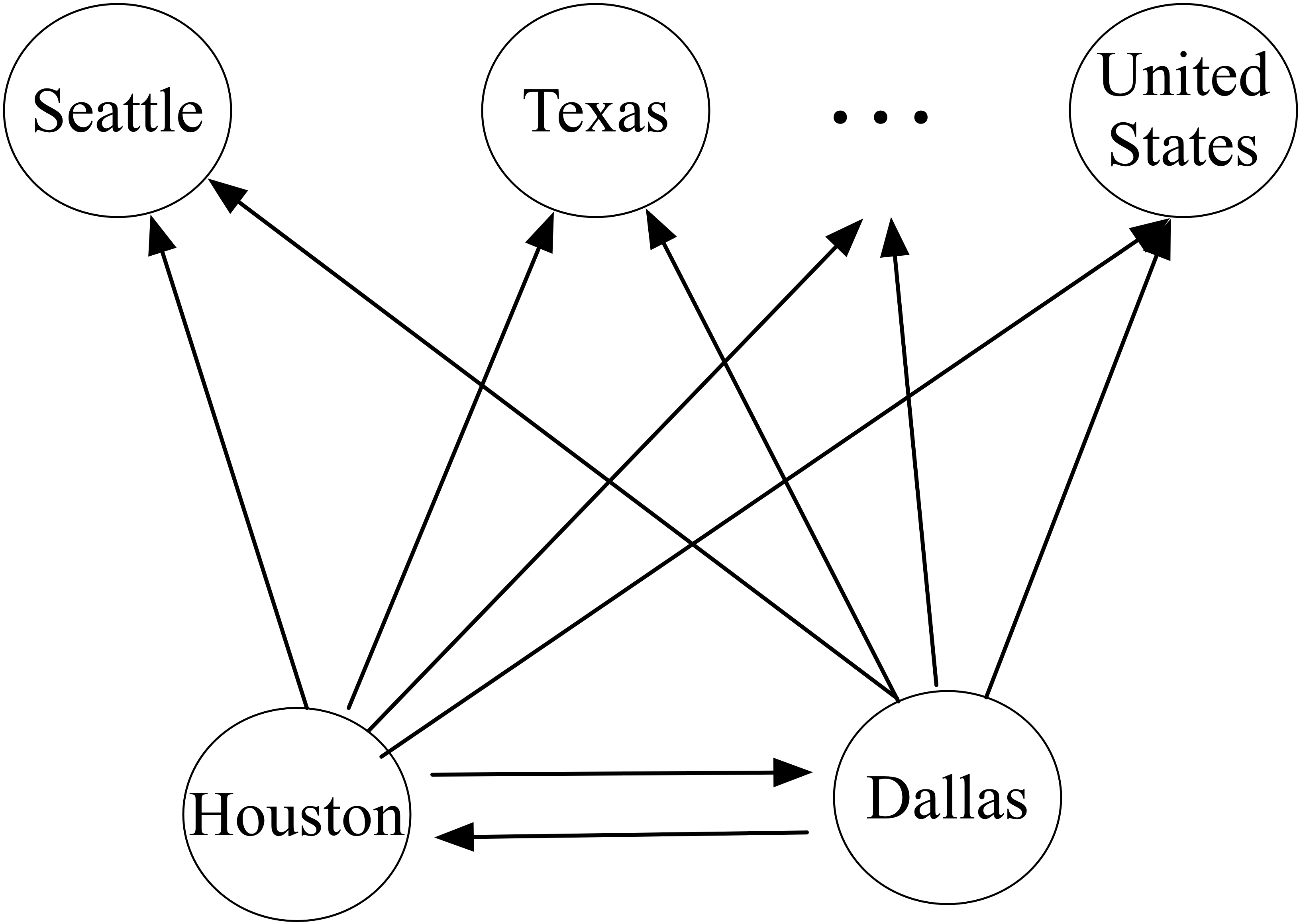}
\caption{The similar topological structure of "Houston" and "Dallas".}
\label{fig:toplogical}
\end{figure}

The entity proximity graph is a graph that captures the semantic relations of the entities. The entities with similar semantic are proximity in the graph, that means they have a similar topological structure in the graph.  For example, as shown in Figure~\ref{fig:toplogical} (to illustrate more clear, we have omitted some unimportant points and edges), there are direct edges between entities "Houston" and "Dallas"
since they are similar in semantic, where the semantic proximity can be simply evaluated by the number of common neighbors between these two entities in the graph. 

To model the implicit mutual relation, we first construct an entity proximity graph based on the Wikipedia corpora. We count the co-occurrence times of each entity pair in Wikipedia corpora, where ``co-occurrence'' refers to two entities appearing in the same sentence. For example, entities ``Obama'' and ``Hawaii'' exist in the same sentence ``Obama was born in Honolulu, Hawaii.'', then the co-occurrence time of ``Obama'' and ``Hawaii'' will increase 1. 

Each entity is a vertex in the entity proximity graph. An edge will be formed if the co-occurrence time of an entity pair is up to a pre-defined threshold. Furthermore, we model the entity proximity graph as a weighted graph, where a weight of each edge is computed as follows:
$$ w_{i,j} = \frac{\log{(co_{i,j})}}{\log{(\max_{k,l}\{co_{k,l}\})}},$$
where the value of $co_{i,j}$ denotes the co-occurrence times of entity pair $(e_i, e_j)$, and $\max_{k,l}\{co_{k,l}\}$ denotes max co-occurrence times of all entity pairs.

In the weighted graph, two vertices with similar topological structure 
indicate that the corresponding entities have similar semantics in unlabeled corpora. Thus, once we construct the entity proximity graph, the implicit mutual relation can be preserved in it.

\subsubsection{Entity embedding learning}
A natural question is how to ensemble the proximity graph, i.e., the implicit mutual relations, into a relation extraction framework. Following the state-of-the-art network embedding approach~\cite{line}, we model the implicit mutual relation of entity pair via learning the vertex embedding in the entity proximity graph.

Our goal is to learn the vertex embedding such that vertices with a similar topological structure in the graph are near neighbors in the low-dimensional space. To preserve the graph structure, we define the \textit{first-order} proximity to capture the observed links in the proximity graph, and define the \textit{second-order} proximity to capture the higher-order proximity between vertices in the proximity graph.

To model the \textit{first-order} proximity, for the edge between entities $e_i$ and $e_j$, the joint probability between $e_i$ and $e_j$ can be defined as follows:
\[P(e_i, e_j) = \frac{1}{1+exp(-\textbf{u}_i^T \cdot \textbf{u}_j)},\]
where $\textbf{u}_i \in R^d$ is the vector representation of entity $e_i$ in the $d$-dimensional space. A superior way to preserve the \textit{first-order} proximity is to minimize the distance between $P(e_i,e_j)$ and its empirical probability. When the KL-divergence is chosen to measure this distance, the objective function is as follows:
\[O_1 = - \sum_{(i,j) \in E} w_{ij}\cdot \log {P(e_i,e_j)}.\]

To model the \textit{second-order} proximity, we assume that vertices with many shared neighbors are similar to each other. For each directed edge $(e_i, e_j)$ in the proximity graph, the probability of ``context'' $e_j$ generated by vertex $e_i$ is defined as
\[P(e_j|e_i) = \frac{\exp{(\textbf{u}_j^T\cdot \textbf{u}_i )}}{\sum_{k=1}^{|V|}{\exp{(\textbf{u}_k^T\cdot \textbf{u}_i )}}},\]
where $|V|$ denotes the amount of vertices.

To preserve the \textit{second-order} proximity, we minimize the distance between $P(e_j|e_i)$ and its empirical probability. Similarly, when the KL-divergence is chosen, the objective function is as follows:
\[O_2 = - \sum_{(i,j) \in E} w_{ij}\cdot \log{P(e_j|e_i)}.\]

In practice, computation of the conditional probability $P(e_j|e_i)$ is extremely expensive. A simple and effective way is to adopt the negative sampling approach mentioned in~\cite{line}. Thus, the above objective function can be simplified to
\[O_2 = \log{\sigma(\textbf{u}_j^T \cdot \textbf{u}_i)} + \sum_{i=1}^K E_{e_n \sim N(e_i)} [\log{\sigma(-\textbf{u}_n^T \cdot \textbf{u}_i)}],\]
where $\sigma(x) = 1/(1 + exp(-x)) $ is the sigmoid function, and $K$ is the number of negative edges. The first term models the observed links, and the second term models the negative links drawn from the noise distribution.

To embed the vertices in the proximity graph, we preserve both the \textit{first-order} proximity and \textit{second-order} proximity separately, then obtain the embedding vector for a vertex by concatenating corresponding embedding vectors learned from the two models.

\subsubsection{Implicit mutual relation}
The vertex embedding vector models the semantic information of an entity. The semantically similar entities, therefore, have close embedding vectors in the embedded space. Thus, we can represent the implicit mutual relation of entities with the entity embedding. The implicit relation between entities $e_i$ and $e_j$ can be represented as follows:

$$MR_{i,j} = \textbf{U}_j - \textbf{U}_i,$$
where $\textbf{U}_i$ is the embedding vector of entity $e_i$.

\subsection{Entity Type Embedding}
In intuition, entity types are beneficial to predict the relation between entities. For example, \textit{/people/person/place\_of\_birth} is the relation between \textit{Location} and \textit{Person}, rather than \textit{Person} and \textit{Person}. Existing works~\cite{P94-100, P1183-1194, reside2018} have also shown that entity type information plays a positive role in relation extraction.

Instances in distance supervision learning are based on the sentences aligned to the knowledge graph, where the entity type information is readily available. Our model uses the entity types defined in FIGER~\cite{P94-100}, which defines 112 fine-grained entity types. To avoid over-parameterization, our model only employs 38 coarse entity types which form the first hierarchy in FIGER. Each entity type is embedded into $k_t$ dimensional space to get the embedding vector of an entity type. When an entity has multiple types, we take the average over the embedding vectors.

We concatenate the embedding of the types for the target entity pair $(e_i, e_j)$ as follows:

$$T_{i,j} = Concat(Type_i,Type_j),$$
where $Type_i$ is the embedding of type for entity $e_i$.

\subsection{Piecewise CNN with Sentence-Level Attention}
The third component of our approach adopts the sentence-level attention to choose high-quality sentences to training our approach. This component consists of three indispensable steps:
\begin{itemize}
  \item[(1)] \textbf{Sentence Embedding}: Each sentence $s_i$ in a training sentences bag $S = \{s_1,s_2,\cdots,s_n\}$ should be represented by word embedding and relative position embedding. Relative position means the relative position of all words in the sentence to the target entities.

    \item[(2)] \textbf{Sentence Encoding}: As the previous works(\cite{pcnn},\cite{senatt}) shown, the convolutional neural networks with piecewise max pooling (PCNN) is a fast and effective way to encode the sentence. Consequently, we get the encoding of each sentence via using PCNN.
    
      \item[(3)] \textbf{Sentence-Level Attention}: The distant supervision learning is suffered from noisy labels, i.e., not all sentences in a bag can express the relation for the targeted entity pair. To address this issue, we utilize the sentence-level attention to mitigate effects from the noise sentence. For the encode of each sentence bag, the model gives each sentence in the bag a score according to the quality of this sentence. 
      The encoding of the $i$th sentence bag can be represented as follows:
       \[X_{bag_i} = \sum_{j\in bag_i} \alpha_j x_j,\]
        where the $X_{bag_i}$ denotes the bag formed by all training sentences of $i$th entity pair. The score $\alpha_j$ for sentence $j$ is calculated by the selective sentence attention. It's defined as:
       
       $$\alpha _ { j } = \frac { \exp \left( q _ { j } \right) } { \sum _ { k } \exp \left( q _ { k } \right) } ,$$
       
       where $q_j$ is a query-based function which scores how well the sentence $j$ and the predict relation $r$ matches. We use the bi-linear function to calculate the scores:
       
       $$q _ { j } = \mathbf { x } _ { j } \mathbf { A } \mathbf { r }, $$
       
       where $\mathbf{A}$ is a weighted diagonal matrix, and $\mathbf{r}$ is the query vector associated with relation $r$.
\end{itemize}

\subsection{Combination of Entity Information and RE Method}

The implicit mutual relation is a semantic relation between entity pairs. Given the targeted relation set $\{r_1,r_2,\cdots,r_m\}$, the entity pairs with similar implicit mutual relation possibly have the same relation in the relations set. Therefore, we can infer the confidence that the target entity pair has relation $r_i$ via using the implicit mutual relation. We use a fully connected layer with a Softmax activation to calculate the confidence score for each relation. For a target entity pair $(e_i,e_j)$, the confidence inferred from the implicit mutual relation is:

$$C_{MR_{i,j}} = Softmax(W_{MR} MR_{i,j} + b_{MR}),$$
where the $W_{MR}$ and $b_{MR}$ are the parameters of the fully connected layer.

Meanwhile, the entity type information can also give a confidence score to $r_i$ according to the entity type constraints of a relation. We concatenate the type embedding of the target entity pair and then use a fully connected layer with a Softmax activation function to calculate the confidence score. As shown below:

$$C_{T_{i,j}} = Softmax(W_T T_{i,j} +b_T ), $$
where the $W_T$ and $b_T$ are the parameters of the fully connected layer. 

The original RE model can give a primary prediction of the probability of each relation:

$$RE_{i,j} = Softmax(W_{RE} X_{bag_i} + b_{RE}), $$

where the $X_{bag_i}$ is the $i$th sentence bag which contains all sentences that the target entity pair $(e_i,e_j)$ co-occurrence in. The $W_{RE}$ and $b_{RE}$ are the parameters of the fully connected layer.

Accordingly, we combine these confidence scores with the original relation extraction (RE) model, to achieve a more accurate result. The probability distribution over $m$ relations between entities $e_i$ and $e_j$ can be computed as follows:
\[P(r_{i,j}) = f(w(\alpha C_{MR_{i,j}} + \beta C_{T_{i,j}} + \gamma RE_{i,j})+b),\]
where $f(x)$ is Softmax function. The $\alpha$, $\beta$ and $\gamma$ are the weight of three components, which can be learned by the RE model itself. 


\subsection{Discussion}

The implicit mutual relation can flexibility combine with various relation extraction models. We integrate the implicit mutual relation with some CNN-based and RNN-based models. As shown in Section \ref{subsection:q2}, these relation extraction models have significantly improved when combined with implicit mutual relations. We think the implicit mutual relations can also have a positive effect on some other advanced methods.

In our solution, the external source we used can easily collect. For the implicit mutual relations, the only external source we used is the unlabeled corpora (i.e., the Wikipedia dump), which can be directly downloaded from the Wikipedia website. For the entity types, the relation extraction is a sub-task for Knowledge Graph (KG) construction, which means the entity types could be obtained from the KG in most cases. Even if the entity types information is missing, using implicit mutual relation alone can also improve the performance, as shown in the section~\ref{subsection:q1}.

\section{EXPERIMENTS}~\label{sec:exp}
We conduct comprehensive experiments to evaluate the performance of our proposed approach by comparing with seven competitors and two variants of our approach on two public datasets. Through the empirical study, we aim at addressing the following research questions:
\begin{itemize}
    \item[RQ1:] How does our proposed approach perform comparing with state-of-the-art relation extraction approaches?
    
    \item[RQ2:] Could the implicit mutual relations and entity types improve the performance of existing relation extraction methods, such as GRU, PCNN, and PCNN + ATT, etc? 
    
    \item[RQ3:] How do the implicit mutual relations affect the relation extraction model?
\end{itemize}

In addition, we conduct a case study, which visually demonstrates the effect of the implicit mutual relations.

\subsection{Experimental Settings}
\subsubsection{Datasets}

\begin{figure*}[htp]
\centering
\subfigure[PR curve on NYT dataset]
{
 \centering
 \begin{minipage}[t]{0.45\linewidth}
 \label{fig:pr_NYT}
 \includegraphics[scale=0.58]{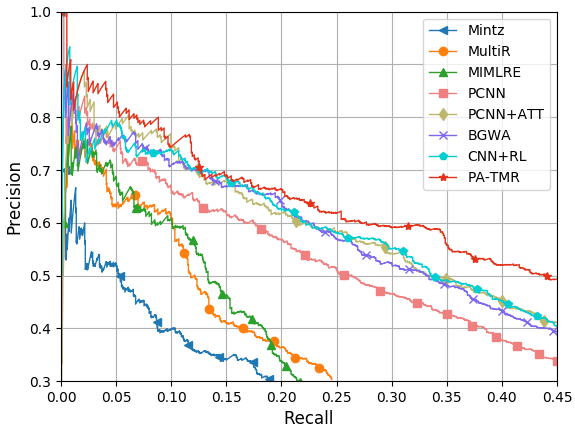}
\end{minipage}
}
\subfigure[PR curve on GDS dataset]
{

 \centering
 \begin{minipage}[t]{0.45\linewidth}
 \label{fig:pr_GDS}
 \includegraphics[scale=0.58]{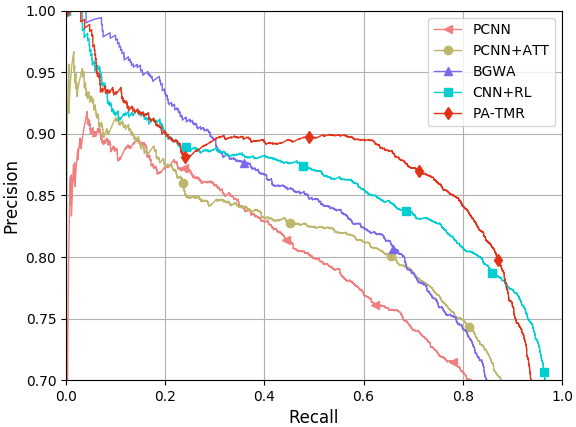}
 \end{minipage}

}
\caption{The Precicion-Recall curve of different algorithms on NYT and GDS datasets}
\label{fig:prcurve}
\end{figure*}

We adopt two widely used public datasets to demonstrate the effectiveness of our method and baselines. They are New York Time(NYT)~\cite{riedel} and Google Distant Supervision (GDS)~\cite{bgwa} datasets, where the statistical descriptions of them are illustrated in Table~\ref{table:dataset}.

\begin{table}[]
\caption{The descriptions of datasets NYT and GDS.}
\centering
\scalebox{1.0}{
\begin{tabular}{|c|c|c|c|c|}
\hline
Datasets & \multicolumn{2}{c|}{\begin{tabular}[c]{@{}c@{}}NYT\\ (\# Relations: 53)\end{tabular}} & \multicolumn{2}{c|}{\begin{tabular}[c]{@{}c@{}}GDS\\ (\# Relations 5)\end{tabular}} \\ \hline \hline
Item     & \# sentences                             & \# entity pairs                             & \# sentences                            & \# entity pairs                            \\ \hline
Training    & 522,611                                 & 281,270                                    & 13,161                                 & 7,580                                     \\ \hline
Testing     & 172,448                                 & 96,678                                     & 5,663                                  & 3,247                                     \\ \hline
\end{tabular}
}

\label{table:dataset}
\end{table}

\begin{itemize}
    \item NYT dataset is generated by annotating entities with Stanford NER tool in the New York Times corpus and then aligns with Freebase to get the relation between entities. The training samples are from the corpus of years 2005-2006, and the testing samples are from the corpus of the year 2007. There are 53 different relations including a relation NA which indicates there is no relation between two entities.

    \item GDS dataset is an extension of the manually annotated data set Google relation extraction corpus. The entities in Google relation extraction corpus are aligned with web documents and then new sentences contain targeted entities are obtained. There are 5 different relations including a relation NA.
\end{itemize}

\subsubsection{Evaluation Metrics}
Similar to most existing works, we evaluate our model with the held-out metrics, which compare the predicting relation facts from the test sentences with those in Freebase. We report the precision, recall, f1-score, precision at top $N$ prediction (P@N), and AUC (area under the Precision-Recall curve). For different threshold, the precision and recall are different, so we report the precision and recall at the point of max f1-score.
In addition, we compute the average score for each metric after running the same experiment five times.

\subsubsection{Parameter Settings}
In the experiment, we use the grid search to tune the optimal model parameters. The grid search approach is used to select the learning rate $\lambda$ for stochastic gradient descent optimizer among \{0.1,0.2,0.3,0.4,0.5\}, the sliding window size $l$ of CNN among \{1,2,3,4,5\}, the number of filters $k$ of CNN among \{180,200,230,250,300\}, and the size of entity type embedding $k_{t}$ among \{10,15,20,25,30,40\}. For the entity embedding size, we follow the setting of~\cite{line}. In Table~\ref{table:para} we show the optimal parameters used in the experiments.

\begin{table}[]
\caption{Parameter settings}
\centering
\begin{tabular}{|c|c|c|}
\hline
Symbol & Description  & Value\\
\hline
$k_{e}$ & Embedding vector size & 128 \\
\hline
$k_{t}$ & Entity type embedding size & 20  \\
\hline
$l$ & Window size                     &3 \\
\hline
$k$ & CNN filters number                 & 230 \\
\hline
$k_p$ & POS embedding dimension           & 5   \\
\hline
$k_w$ & Word embedding dimension          &50   \\
\hline
$lr$ & Learning rate                      & 0.3 \\
\hline
$l$  & Sentence max length                & 120 \\
\hline
$p$  & Dropout probability                 & 0.5 \\
\hline
$n$  & Batch size                         & 160 \\
\hline
\end{tabular}
\label{table:para}
\end{table}

\begin{table*}[]
\caption{Performance Comparison}
\label{table:result}
\centering
\begin{tabular}{|c|c|c|c|c|c|c|c|}
\hline
\textbf{Dataset}     & \textbf{Method} & \textbf{AUC}    & \textbf{Precision} & \textbf{Recall} & \textbf{F1-Score} & \textbf{P@100} & \textbf{P@200} \\ \hline \hline
\multirow{7}{*}{NYT} & PCNN             & 0.3296          & 0.3830             & 0.4020          & 0.3923            & 0.77           & 0.72           \\ \cline{2-8} 
                     & PCNN+ATT         & 0.3424          & 0.3588             & 0.4564          & 0.4018            & 0.75           & 0.75           \\ \cline{2-8} 
                     & BGWA             & 0.3670          & 0.3994             & 0.4451          & 0.4210            & 0.76           & 0.74           \\ \cline{2-8} 
                     & CNN+RL           & 0.3735          & 0.4201             & 0.4389          & 0.4293            & 0.79           & 0.73           \\   \cline{2-8} 
                     & PA-T             & 0.3572          & 0.3779             & 0.4586          & 0.4143            & 0.78           & 0.72           \\ \cline{2-8} 
                     & PA-MR            & 0.3635          & 0.4091             & 0.4410          & 0.4244            & 0.79           & 0.78           \\ \cline{2-8} 
                     & PA-TMR           & \textbf{0.3939} & \textbf{0.4320}    & \textbf{0.4615} & \textbf{0.4463}   & \textbf{0.83}  & \textbf{0.79}  \\ \hline \hline
\multirow{7}{*}{GDS} & PCNN             & 0.7798          & 0.6804             & 0.8673          & 0.7626            & 0.88           & 0.90           \\ \cline{2-8} 
                     & PCNN+ATT         & 0.8034          & 0.7250             & 0.8474          & 0.7814            & 0.94           & 0.93           \\ \cline{2-8} 
                     & BGWA             & 0.8148          & 0.7725             & 0.7162          & 0.8385            & 0.99           & 0.98           \\ \cline{2-8} 
                     & CNN+RL           & 0.8554          & 0.7680             & \textbf{0.9132} & 0.8343            & 1.0            & 0.96           \\ \cline{2-8}
                     & PA-T             & 0.8512          & 0.7925             & 0.8969          & 0.8414            & 0.96           & 0.94           \\ \cline{2-8} 
                     & PA-MR            & 0.8571          & 0.8011             & 0.8947          & \textbf{0.8453}   & 0.97           & 0.94           \\ \cline{2-8}  
                     & PA-TMR           & \textbf{0.8646} & \textbf{0.8058}    & 0.8641          & 0.8339            & \textbf{1.0}   & \textbf{0.98}  \\ \hline
\end{tabular}
\end{table*}

\subsubsection{Baselines}
For evaluating our proposed model, we compare with the following baselines:

\textbf{Mintz}\cite{ds} is a traditional distant supervision model which utilizes multi-class logistic regression to extract relations between entities.

\textbf{MultiR}\cite{hoffmann} utilizes multi-instance learning to combat the noise from distant supervision learning. It introduces a probabilistic graphical model of multi-instance learning which handles overlapping relations.
 
\textbf{MIMLRE}\cite{surdeanu} proposes a graphical model which can jointly model the multiple instances and multiple relations.

\textbf{BGWA}\cite{bgwa} is a bidirectional GRU based relation extraction model. It focuses on reducing the noise from distant supervision learning by using a hierarchical attention mechanism.

\textbf{PCNN}\cite{pcnn} is a CNN based relation extraction model which utilizes the piecewise max pooling to replace the single max pooling to capture the structural information between two entities.

\textbf{PCNN+ATT}\cite{senatt} combines the selective attention over instances with PCNN. The selective attention mechanism is expected to dynamically reduce the weights of those noisy instances, thereby reducing the influence of wrong labeled instances.

\textbf{CNN+RL}\cite{reinforce} contains two modules: an instance selector and a relation classifier. The instance selector chooses high-quality sentences with reinforcement learning. The relation classifier makes a prediction by the chosen sentences and provides rewards to the instance selector.

Based on the state-of-the-art relation extraction approach, PCNN+ATT, \textbf{PA-TMR} is our proposed approach which integrates entity types and implicit mutual relations into PCNN+ATT approach. In addition, we propose two variants \textbf{PA-T} and \textbf{PA-MR} which only adopt entity type and implicit mutual relation to improve PCNN+ATT approach, respectively.

\subsection{Performance Comparison (\textbf{RQ1})}
\label{subsection:q1}

To verify the effectiveness of our model, we compare our PA-TMR model with baselines on both NYT and GDS datasets as demonstrated in Figure~\ref{fig:prcurve}. We use the results reported in Lin et al.~\cite{senatt} for the performances of non-neural baselines Mintz~\cite{ds}, MultiR~\cite{hoffmann} and MIMLRE~\cite{surdeanu} on NYT dataset. As shown in Figure~\ref{fig:pr_NYT}, all the non-neural baselines obviously worse than the neural baselines, so we only report the results of neural baselines on GDS dataset. As illustrated in Table~\ref{table:result} and Figure~\ref{fig:prcurve}, we have the following key observations:

\begin{itemize}
    \item The performance of PCNN is worse than the other neural models. This is due to the factor that the PCNN model does not improve to alleviate the impact from the noisy training sentences, while other neural baselines and our PA-TMR method utilize some techniques, such as reinforcement learning or attention mechanism, to deal with the problem of noisy training sentences. Meanwhile, it reveals the practical necessity to deal with the problem of noisy training sentences in our PA-TMR method.
    
    \item Our PA-TMR model not only outperforms all the neural baselines significantly, but also has more obvious advantage when the recall increases as demonstrated in Figure~\ref{fig:prcurve}. This is due the factors that: (1) all the neural baselines only employ the training corpus to extract relations; (2) the noisy training sentences in distant supervision corpora exacerbates insufficient training problem for the RE models. However, PA-TMR combines the implicit mutual relations and entity types to improve the neural relation extraction. This points to the positive effect of integrating both the implicit mutual relations and entity types into the RE model.
    
    
    \item Comparing to the variants of our PA-TMR method, both PA-T and PA-MR outperform the basic model PCNN+ATT. This improvement illustrates that both the implicit mutual relations and entity types have the positive effect on extracting relations again. Furthermore, PA-TMR achieves the best performance compared to its variants. This sheds the light on the benefit of the interaction of the implicit mutual relations and entity types.
\end{itemize}

\begin{figure}[h]
\centering

\begin{minipage}[t]{0.42\linewidth}
\centering
\subfigure[NYT dataset]
{
    \label{fig:improve-nyt}
    \includegraphics[scale=0.26]{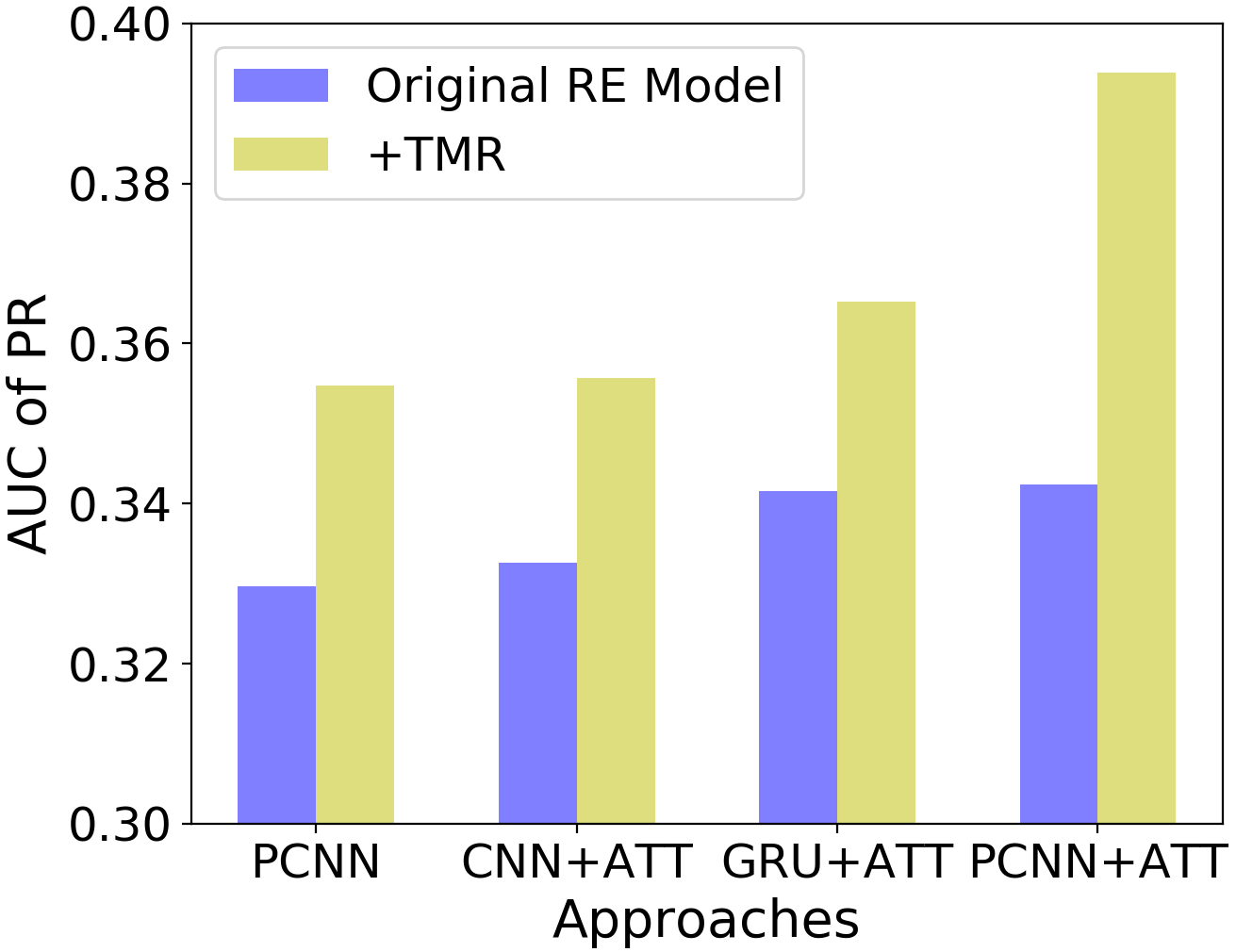}}

\end{minipage}
\hspace*{\fill}
\centering
\begin{minipage}[t]{0.42\linewidth}
\centering
\subfigure[GDS dataset]
{
    \label{fig:improve-gds}
    \includegraphics[scale=0.26]{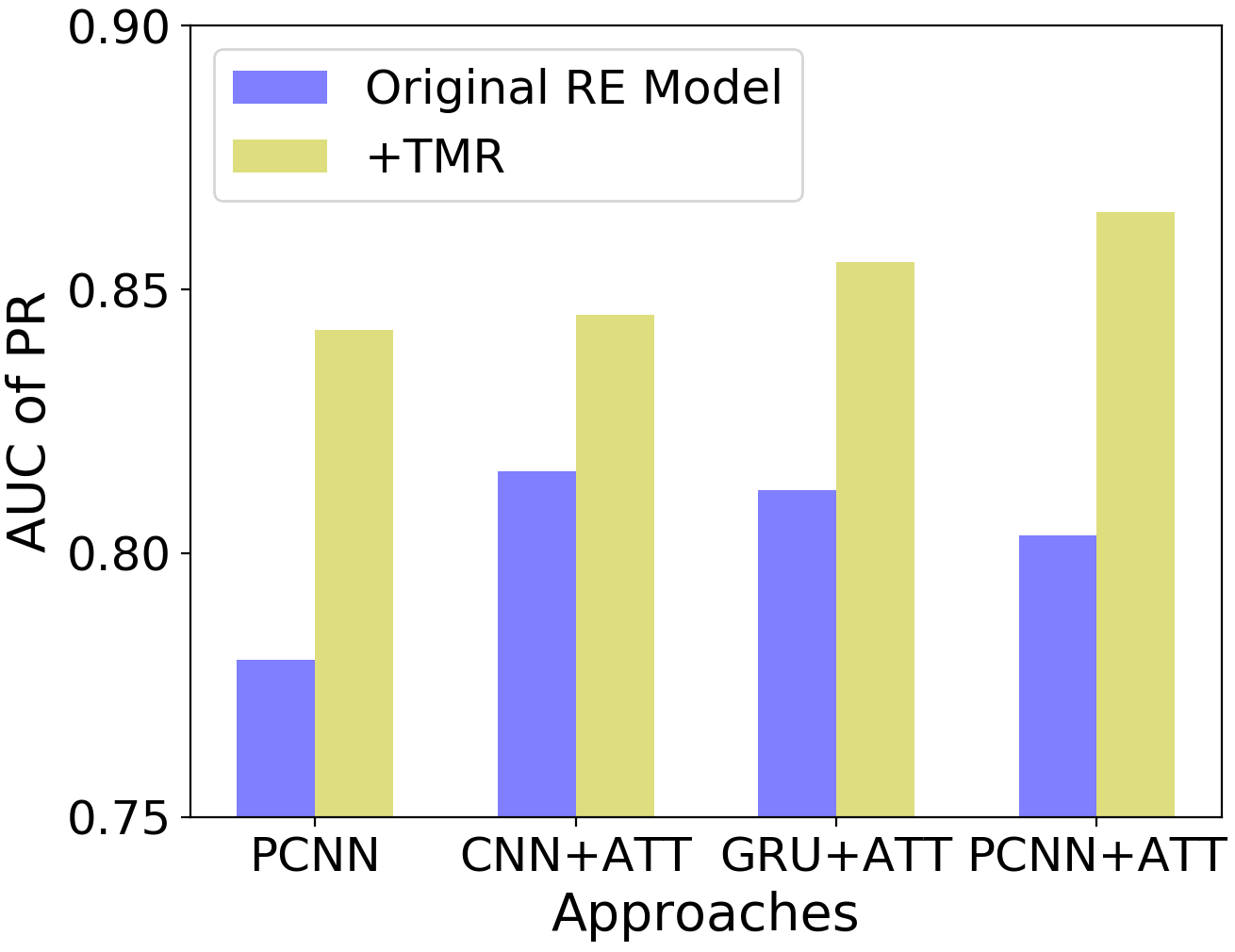}
}
\end{minipage}
\hspace*{\fill}
\caption{The flexibility of our proposed neural RE framework and the improvement of the implicit mutual relations and entity types}
\label{fig:improve}
\end{figure}

\begin{figure*}[ht]
\centering

\begin{minipage}[t]{0.45\linewidth}
\centering
\subfigure[NYT dataset]
{
    \label{fig:f1_dif_nyt}
    \includegraphics[scale=0.33]{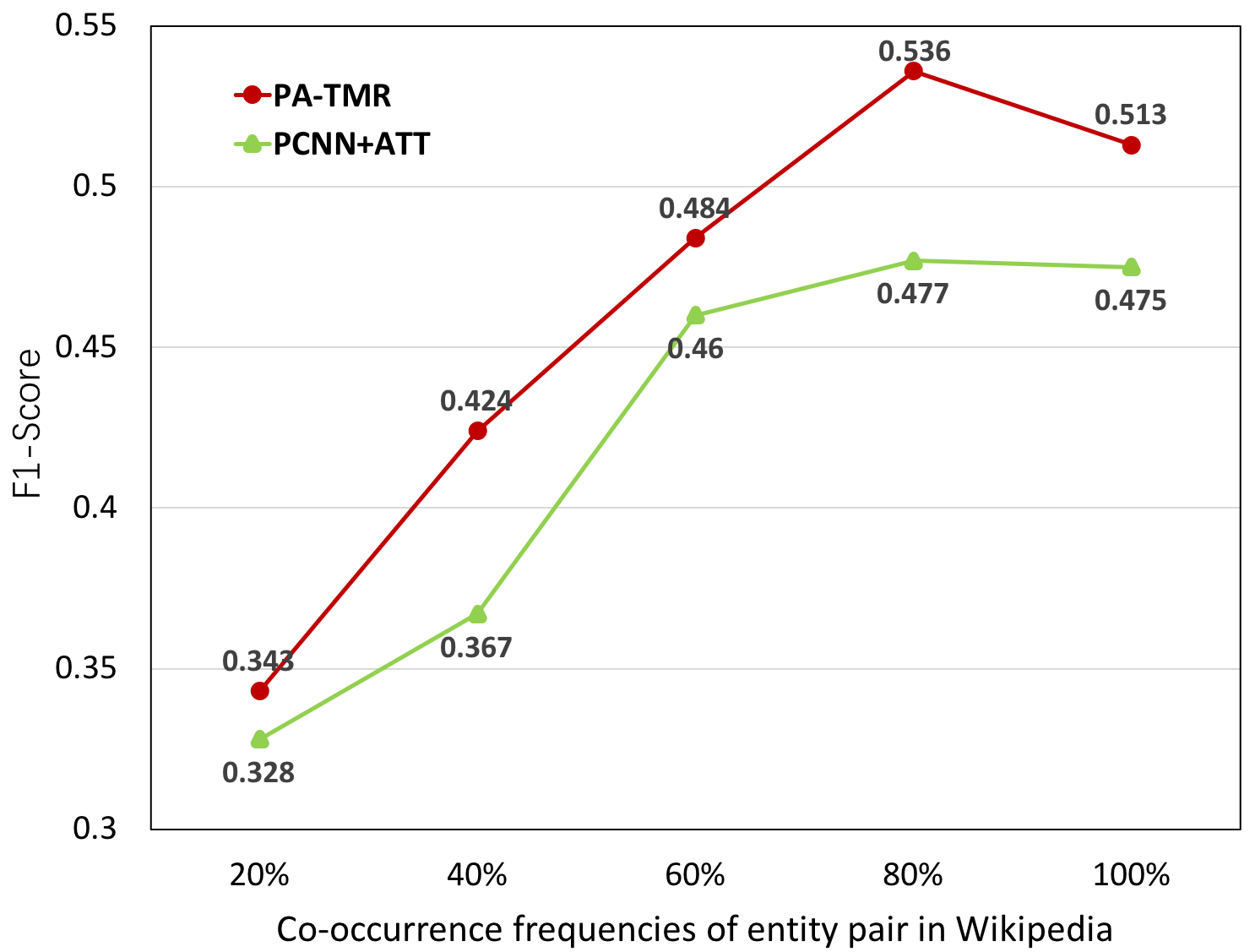}
}

\end{minipage}
\hspace*{\fill}
\centering
\begin{minipage}[t]{0.45\linewidth}
\centering
\subfigure[GDS dataset]
{
    \label{fig:f1_dif_gids}
    \includegraphics[scale=0.33]{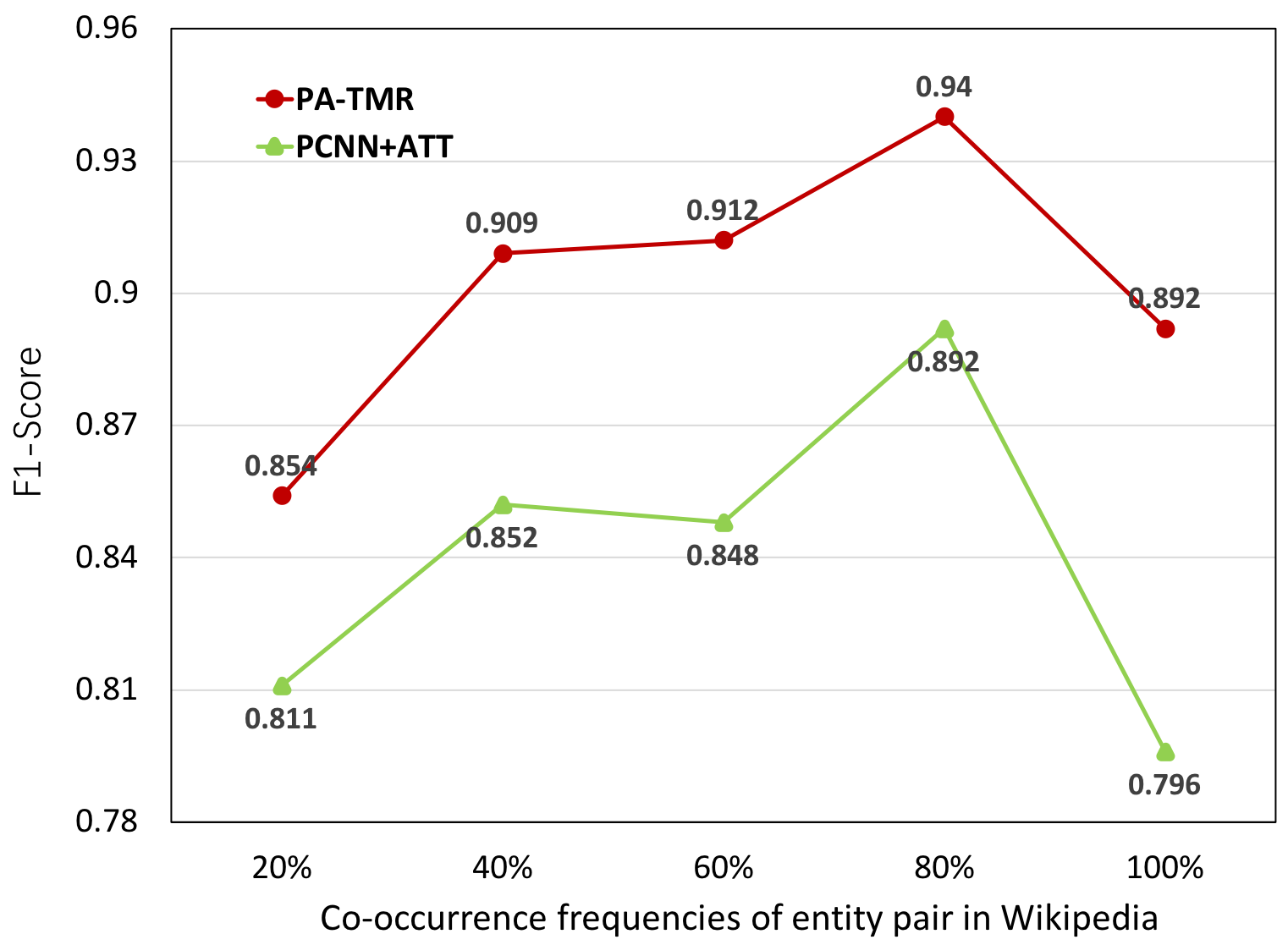}
}
\end{minipage}
\hspace*{\fill}
\caption{The f1-score of the test sets with different co-occurrence frequencies of entity pairs.}
\label{fig:co_occur}
\end{figure*}

  

  
  

\subsection{Flexibility of Our Method (\textbf{RQ2})}
\label{subsection:q2}
To illustrate the flexibility of our PA-TMR method, we corporate the components of implicit mutual relations and entity types into the other neural relation extraction approaches, such as GRU based model with sentence level attention (GRU+ATT), CNN + ATT ~\cite{senatt}, PCNN ~\cite{pcnn}, and PCNN + ATT ~\cite{senatt}. As elaborated in Figure~\ref{fig:improve}, we have the following key observations:

\begin{itemize}
    \item Comparing to the original models, each improved model achieves 2\%-7\% improvement by combining the implicit mutual relations, entity types, and distant supervision training corpora. The better performance of the improved models is twofold: (1) all entity pairs with the similar semantic are helpful to extract relation for the target entity pair; (2) the implicit mutual relations can further alleviate the impact from the noisy data in the training corpora. As such, it reveals that only distant supervision training corpora is insufficient for predicting the relation of the target entity pair.
    
    \item The original RE models are CNN-based (CNN + ATT, PCNN, PCNN + ATT) or RNN-based (GRU+ATT) approaches. The experimental result illustrates that the basic CNN-based and RNN-based models can achieve significant improvement via only integrating our implicit mutual relations into them without any modification of original approaches. This sheds light on the flexibility of using our proposed implicit mutual relations. Meanwhile, it indicates that the implicit mutual relations can be integrated into most of neural relation extraction methods easily. 
\end{itemize}

\subsection{The Effect of Implicit Mutual Relations (\textbf{RQ3})}

\begin{figure*}[h]
\centering

\begin{minipage}[t]{0.45\linewidth}
\centering
\subfigure[NYT dataset]
{
    \label{fig:infreq_nyt}
    \includegraphics[scale=0.57]{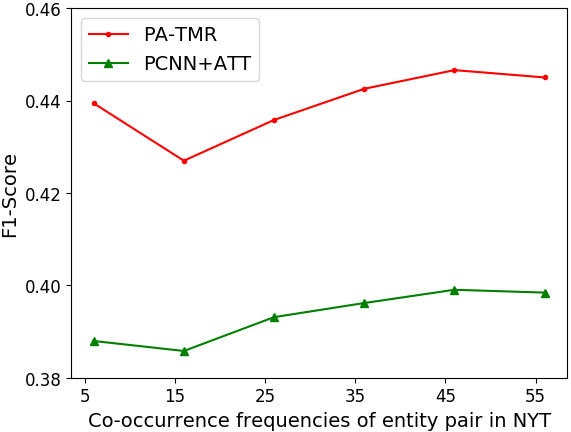}
}

\end{minipage}
\hspace*{\fill}
\centering
\begin{minipage}[t]{0.45\linewidth}
\centering
\subfigure[GDS dataset]
{
    \label{fig:infreq_gds}
    \includegraphics[scale=0.57]{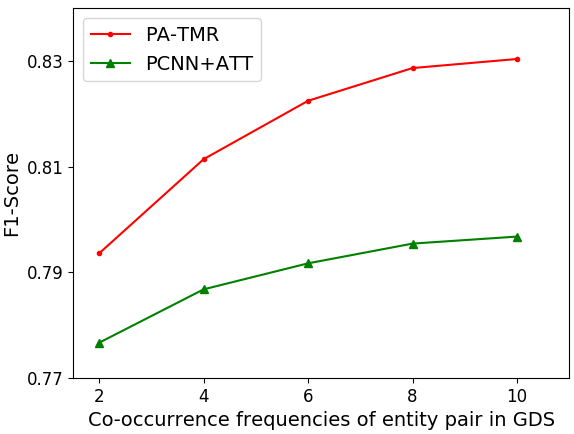}
}
\end{minipage}
\hspace*{\fill}
\caption{The f1-score of the entity pairs with the different co-occurrence frequencies in original dataset.}
\label{fig:infreq}
\end{figure*}

To illustrate the effectiveness of integrating the implicit mutual relations, we first evaluate the performance of our PA-TMR method with different co-occurrence frequencies in unlabeled corpora as illustrated in Figure~\ref{fig:co_occur}, from which we can find the positive effect of implicit mutual relations of entity pairs with different co-occurrences frequencies in the unlabeled corpora. Then we demonstrate the performance of PA-TMR considering entity pairs with infrequent training instances in the training corpora as shown in Figure~\ref{fig:infreq}, which verifies the positive effect of implicit mutual relations for infrequent entity pairs.

\subsubsection{Improvement from implicit mutual relations}

As illustrated in Figure~\ref{fig:co_occur}, we sort the entity pairs by their co-occurrence frequencies in unlabeled corpora (Wikipedia) and then evaluate the performance for the entity pairs with different co-occurrence frequencies, where the x-axis denotes the quantile of co-occurrence frequencies of entity pairs in Wikipedia, and the y-axis denotes the corresponding F1-score. We have the following key observations:
\begin{itemize}
    \item As the co-occurrence frequencies of entity pairs increase, the F1-score demonstrates an upwards synchronous trend. It reveals that no matter frequent or infrequent co-occurrences of entity pairs in the unlabeled corpora are helpful for improving the performance of our PA-TMR model. This points to the positive effect of all implicit mutual relations collected from the unlabeled corpora. Meanwhile, the implicit mutual relations, which capture the semantic information of both the target entity pair and the entity pairs with similar semantic, contributes to predict relations for the target entity pair;
    
    \item The improvement on the small dataset GDS is much larger than that on NYT dataset. This is due to the factor that: (1) we insufficiently train the original RE model in the smaller dataset; (2) noisy data in a smaller training dataset exacerbates the inadequate issue of training process by utilizing the attention mechanism. The better improvement illustrates that the implicit mutual relations can alleviate the negative impact of insufficient training corpora.
\end{itemize}

\subsubsection{The effect on inadequate training sentences}

As illustrated in Figure~\ref{fig:infreq}, we evaluate the impact of inadequate training sentences, where the x-axis denotes the \# training sentences in the distant supervision training corpora, and the y-axis denotes the F1-score of relation extraction for the entity pairs with fixed number of training sentences. We have the following key observations:
\begin{itemize}
    \item The performance of original PCNN + ATT increases as an entity pair has more training sentences in the distant supervision training corpora. It reveals that inadequate training sentences have negative impact on extracting relations.

    \item Our PA-TMR method outperforms the PCNN+ATT for extracting relations for the entity pairs with inadequate training sentences significantly. This is due to the factor that our mined implicit mutual relations contribute to predict the relations of entity pairs with inadequate training sentences.
\end{itemize}

\subsection{Case Study}
In the above experiment, we have identified the effect of implicit mutual relation of entity pairs in extracting relations. It is a natural question that how the improving mechanism of the implicit mutual relations works in the extracting process. 
Note that the implicit mutual relation is represented as the entity embedding learned from the entity proximity graph. 
Therefore, we conduct a case study to demonstrate the meanings of the implicit mutual relation after entity embedding.

\begin{figure}[htp]
\centering
\includegraphics[scale=0.43]{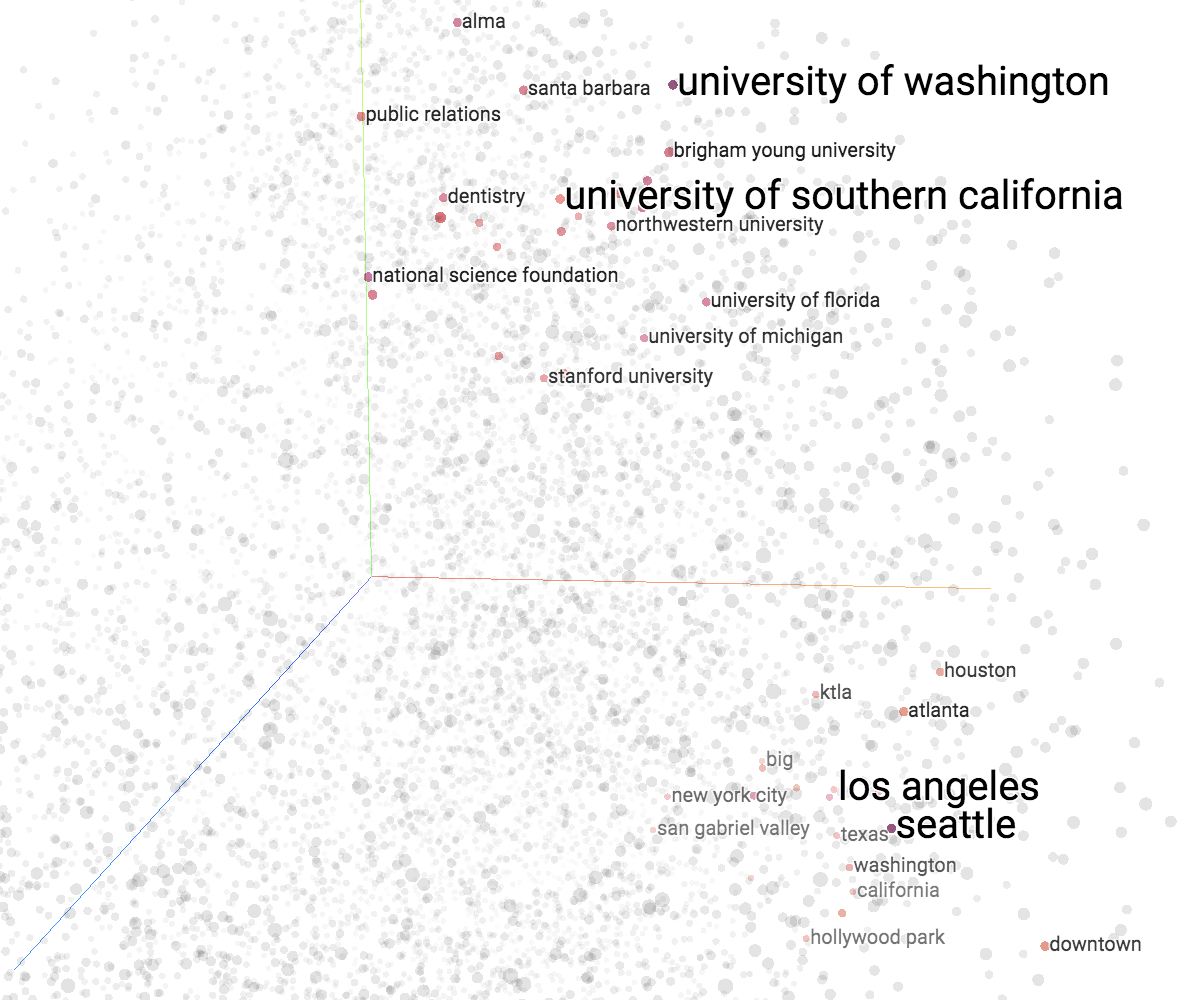}
\caption{Visualization of the entity embedding learned from the entity proximity graph.}
\label{fig:entityVisual}
\end{figure}

As shown in Figure~\ref{fig:entityVisual}\footnote[1]{This picture is produced by the Embedding Projector \url{https://projector.tensorflow.org/}}, the embedding vectors of entities are projected into 3D space. We show the nearest entities of \textbf{Seattle} and \textbf{University of Washington} in the figure. We can observe that most of the nearest entities of \textbf{Seattle} are cities in the USA, and most of the nearest entities of \textbf{University of Washington} are universities. This suggests that the entities with similar semantic would be closed in the embedding space. The top 10 nearest entities of Seattle and University of Washington are shown in table \ref{table:emb}. However, there are some entities whose semantics are not close to the target entity pair, such as "San Gabriel Valley". Therefore, in future work, we can adopt more advanced methods to learn the representation of entities to alleviate this problem.

For entity pair (\textbf{University of Washington}, \textbf{Seattle}), its implicit mutual relation is similar with many other entity pairs, such as (\textbf{University of Southern California}, \textbf{Los Angeles}) and (\textbf{Stanford University}, \textbf{California}), etc. Thus, our model tends to correctly predict the relation between ``Seattle'' and ``University of Washington'' if we have the high quality training instances for entity pairs (\textbf{University of Southern California}, \textbf{Los Angeles}) and ( \textbf{Stanford University}, \textbf{California}), or the RE approaches correctly predict the relations between entity pairs (\textbf{University of Southern California}, \textbf{Los Angeles}) and (\textbf{Stanford University}, \textbf{California}).

\begin{table}[]
\centering
\caption{The Nearest Entities of Seattle and University of Washington in Embedding Space}
\label{table:emb}
\begin{tabular}{|c|c|c|}
\hline
Top N & \textbf{University of Washington} & \textbf{Seattle} \\ \hline \hline
1     & University of Florida             & New York City    \\ \hline
2     & University of South California    & Washinton        \\ \hline
3     & Brigham University                & California       \\ \hline
4     & Stanford University               & Los Angeles      \\ \hline
5     & Northwestern University           & Texas            \\ \hline
6     & Ohio State University           & Houston            \\ \hline
7     & University of Michigan            & Downtown            \\ \hline
8     & Bowling Green             & San Gabriel Valley            \\ \hline
9     & Alma           & Atlanta            \\ \hline

10     & University of Kentucky           & Cleveland            \\ \hline

\end{tabular}
\end{table}

\section{CONCLUSION and FUTURE WORK}
~\label{sec:conculsion}
We have presented a unified approach for improving the existing neural relation extraction approaches. In contrast to the existing neural RE models that train the model by only using the distant supervision training corpora, We learn the implicit mutual relations of entity pairs from the unlabeled corpora via embedding the vertices in the entity proximity graph into a low-dimensional space. Meanwhile, our proposed implicit mutual relations are easily and flexibly integrated into existing relation extraction approaches. The experimental results outperform state-of-the-art relation extraction approaches, and manifest that the implicit mutual relations of entity pairs and the entity type information have a positive effect for relation extraction.

In this work, we have only employed the first-order and second-order proximity to capture the implicit mutual relations when we learn the vertex embedding in the entity proximity graph. Thus, it may fail for vertices that have few or even no edges. To address this issue we plan to utilize the graph neural networks (GNNs)~\cite{gcn} or Graph Attention Networks (GATs)~\cite{gat} to model auxiliary side information, such as numerical features and textual descriptions.
In addition, we adopt sentence-level attention to mitigate effects from the noise sentence.
As mentioned by Y. Liu et al.~\cite{senatt}, the attention mechanism usually alleviates the negative impact of noisy training data. Lastly, we are interested in integrating the other attention mechanism to alleviate the problem.

\bibliographystyle{IEEEtran} 
\bibliography{reference}
\vspace{12pt}

\end{document}